\documentclass[letterpaper]{article} % DO NOT CHANGE THIS
\usepackage{aaai25}  % DO NOT CHANGE THIS
\usepackage{times}  % DO NOT CHANGE THIS
\usepackage{helvet}  % DO NOT CHANGE THIS
\usepackage{courier}  % DO NOT CHANGE THIS
\usepackage[hyphens]{url}  % DO NOT CHANGE THIS
\usepackage{graphicx} % DO NOT CHANGE THIS
\usepackage{natbib}  % DO NOT CHANGE THIS AND DO NOT ADD ANY OPTIONS TO IT
\usepackage{caption} % DO NOT CHANGE THIS AND DO NOT ADD ANY OPTIONS TO IT
\usepackage{algorithm}
\usepackage{algorithmic}
\usepackage{amsmath}
\usepackage{amssymb}

\def \ie {\emph{i.e.}~}
\def \eg {\emph{e.g.}~}
\def \etc {\emph{etc.}~}
\def \etal {\emph{et al.}~}

\usepackage{newfloat}
\usepackage{listings}
\DeclareCaptionStyle{ruled}{labelfont=normalfont,labelsep=colon,strut=off} % DO NOT CHANGE THIS
\floatstyle{ruled}
\newfloat{listing}{tb}{lst}{}
\floatname{listing}{Listing}
\title{Enhancing Generalized Few-Shot Semantic Segmentation via\\ Effective Knowledge Transfer}
\author{
    Xinyue Chen\textsuperscript{\rm 1},
    Miaojing Shi\textsuperscript{\rm 2}\thanks{Corresponding author.},
    Zijian Zhou\textsuperscript{\rm 1},
    Lianghua He\textsuperscript{\rm 2},
    Sophia Tsoka\textsuperscript{\rm 1}
}
\affiliations{
    \textsuperscript{\rm 1}Department of Informatics, King's College London, UK\\
    \textsuperscript{\rm 2} College of Electronic and Information Engineering, Tongji University, China\\
    \{xinyue.1.chen, zijian.zhou, sophia.tsoka\}@kcl.ac.uk,
    \{mshi, helianghua\}@tongji.edu.cn
}

\begin{document}

\maketitle

\begin{abstract}
Generalized few-shot semantic segmentation (GFSS) aims to segment objects of both base and novel classes, using sufficient samples of base classes and few samples of novel classes. Representative GFSS approaches typically employ a two-phase training scheme, involving base class pre-training followed by novel class fine-tuning, to learn the classifiers for base and novel classes respectively. Nevertheless, distribution gap exists between base and novel classes in this process. To narrow this gap, we exploit effective knowledge transfer from base to novel classes. First, a novel prototype modulation module is designed to modulate novel class prototypes by exploiting the correlations between base and novel classes. Second, a novel classifier calibration module is proposed to calibrate the weight distribution of the novel classifier according to that of the base classifier. Furthermore, existing GFSS approaches suffer from a lack of contextual information for novel classes due to their limited samples, we thereby introduce a context consistency learning scheme to transfer the contextual knowledge from base to novel classes. Extensive experiments on PASCAL-5$^i$ and COCO-20$^i$ demonstrate that our approach significantly enhances the state of the art in the GFSS setting. The code is available at: https://github.com/HHHHedy/GFSS-EKT.
\end{abstract}


\section{Introduction} \label{sec:intro}

Semantic segmentation is a fundamental computer vision task with widespread applications in fields like robotics and medical imaging. The advent of fully convolutional network (FCN) and vision transformer (ViT) has led to significant achievements in semantic segmentation~\cite{yu2015multi,2017pspnet,li2022contextual,zhang2022mm-former}. However, supervised learning for this task typically demands a large amount of annotated data yet the trained models cannot recognize novel classes. To mitigate this, few-shot semantic segmentation (FSS) has been proposed to develop models that can effectively segment objects of novel classes using only a handful of annotated support samples. Recently, FSS methods \cite{Wang_2019_ICCV,lang2023base,liu2020prototype,li2021asgnet,chen2024memory} have achieved significant progress; but they are constrained to segment only objects of novel classes while ignoring base classes. To overcome this limitation, generalized few-shot semantic segmentation (GFSS) extends FSS to segment both base and novel classes. 

\begin{figure}[t]
	\centering
	\includegraphics[width=1.0\columnwidth]{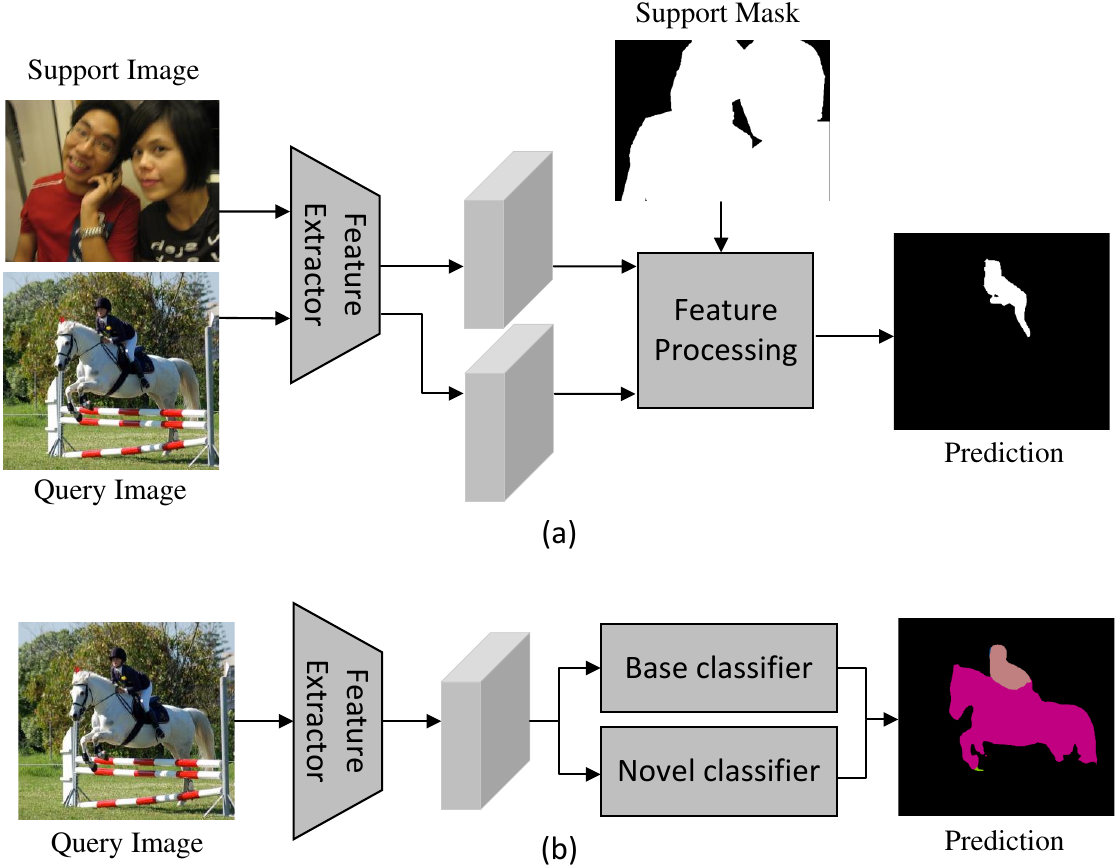}
    %\vspace{-6mm}
	\caption{Illustration of (a) FSS and (b) GFSS: FSS models predict only the novel class specified by the support image, whereas GFSS models can predict both base and novel classes at the same time. During inference, GFSS models do not rely on support images of novel classes any more, as they are fine-tuned using all samples of novel classes to form a novel classifier. In this context, ``horse'' is a base class, while ``person'' represents a novel class.}
        %\vspace{-2mm}
	\label{fig:general}
\end{figure}

%\vspace{-4mm}

Representative GFSS approaches \cite{tian2022generalized,liu2023learning,Hajimiri_2023_CVPR,lu2023prediction,zhang2024memory} adopt a two-phase training process: in the pre-training phase, the model is trained on sufficient base samples to segment objects of base classes; afterward, in the fine-tuning phase, the model is updated with a few annotated samples of novel classes, enabling simultaneous segmentation of both base and novel classes. Figure \ref{fig:general} illustrates the comparison between FSS and GFSS. This two-phase training approach often results in poor performance on novel classes due to the distribution gap between base and novel classes. This distribution gap causes the feature extractor and base classifier, which are pre-trained on abundant samples of base classes, to fail in effectively capturing the features of novel classes and successfully segmenting novel classes within images. A typical solution is to fine-tune the feature extractor and base classifier using the limited novel samples available; however, this can lead to performance degradation on the base classes. While training a separate novel classifier is possible, achieving balanced base and novel classifiers simultaneously remain challenging. %Furthermore, the feature extractor and other components that are trained with limited samples of novel classes can not be reliable.
%Considering that base and novel classes often share correlations, the essential idea of this paper is to leverage these correlations to enable effective knowledge transfer from base to novel classes.

%This issue arises because some components, such as feature representations and the classifier for novel classes, are trained on few novel samples.
%1) the feature extractor trained on only base classes fails to capture optimal features for novel classes due to a feature distribution shift between base and novel classes, which obstructs model generalization; 2) the weight distribution of the learned novel classifier is neither representative nor reliable due to the limited samples of novel classes, whereas the base classifier is well-learned with ample samples. 
%resulting in1) improper establishment of the mean and vanriance of the weight distribution of novel classes with few support samples compared to those of base classes;
%1) a strong bias towards novel classes when fine-tuning a pretrained model with only novel samples, even when the feature extractor is frozen, which leads to the misclassification of base class pixels as novel class pixels. 
%2) a division between base and novel prototypes, impeding the recognition of novel prototypes by the model.
%2) the challenge of obtaining representative and unique novel prototypes given limited support samples, especially when requiring them to be orthogonal.
%With few samples for novel classes, enforcing orthogonality not only fails to learn the unique features of these classes but also leads to biased learning.
In this work, we propose an Effective Knowledge Transfer approach for enhancing GFSS, namely GFSS-EKT, to prevent performance degradation on base classes and improve performance on novel classes. Our method employs a two-phase training scheme. In the pre-training phase, we extract features from input base class samples using a feature extractor and decompose them by projecting them onto learnable base prototypes representing different classes. Then, a base classifier is trained to perform a classification on base classes. The same feature extraction is used in the fine-tuning phase, the features of novel class samples are decomposed by both base and novel prototypes. A novel classifier is then learned alongside the base classifier to classify both base and novel classes. We make three main contributions. First, we propose a \emph{novel prototype modulation} module that adjusts novel class prototypes by leveraging their correlations with base class prototypes. Second, we introduce a \emph{novel classifier calibration} module that adjusts the weight distribution of the novel classifier through mean shifting and standard deviation scaling, obtaining a more reliable novel classifier. Finally, current GFSS approaches face the challenge of lacking contextual information for novel classes when fine-tuning models with few novel samples. To tackle it, we incorporate base samples into the fine-tuning phase and introduce a \emph{context consistency learning} scheme between two augmented versions of a given base sample. The augmentation operations are carefully designed to focus on the meaningful contextual part of the base image so that the relevant knowledge can be transferred from the base to novel classes. 

We evaluate our method on two public datasets, PASCAL-5$^i$ and COCO-20$^i$. The experimental results demonstrate that our method achieves significant improvement over the state of the art. 

% \textcolor{red}{Overall, our main contributions can be summarized as:
% \begin{enumerate}
%     \item We propose a novel classifier calibration method that aligns two classifiers by calibrating the weight distribution of the novel classifier to match that of the base classifier. 
%     \item We propose a novel prototype modulation method that exploits the correlative information between base and novel classes to improve novel prototypes.
%     \item We implement a context consistency learning strategy to transfer rich contextual and semantic information from base to novel classes, effectively reducing overfitting on novel classes.
%     \item Extensive experiments on two standard datasets, namely PASCAL-5$^i$ and COCO-20$^i$, demonstrate that our method outperforms the SOTA, with improvements on both base and novel classes.
% \end{enumerate}}

\section{Related Work} \label{sec: related work}

\paragraph{Semantic Segmentation.}

Semantic segmentation is to classify each pixel of the image into specific class. Since the emergence of fully convolutional network (FCN) \cite{long2015fcn}, semantic segmentation has achieved remarkable progress thanks to various advanced techniques. For instance, dilated convolution \cite{yu2015multi}, pyramid pooling module \cite{2017pspnet}, atrous spatial pyramid pooling \cite{chen2017deeplab}, \etc have been proposed to handle varying object sizes in images. Besides, some transformer-based FSS models \cite{hu2022suppressing,lu2021simpler,zhang2021cyc,zhang2022mm-former} have also been proposed since the advent of ViT \cite{dosovitskiy2020vit} in computer vision. For example, MM-Former \cite{zhang2022mm-former} employs Mask2Former \cite{2022mask2former} to generate multiple mask proposals for the query image. Despite the success of these methods, they generally require a large amount of training data to achieve good performance. 

\paragraph{Few-Shot Semantic Segmentation.}

Few-shot semantic segmentation (FSS) aims to segment novel classes given a few support samples, which alleviates the dependence of the segmentation model on a large amount of training data. Previous FSS approaches \cite{Wang_2019_ICCV,liu2020prototype,lang2023base,li2021asgnet,chen2023distilling} typically employ the episodic learning that organizes the base data into multiple episodes, each of which consists of a query image and few support images of the same base class to simulate the few-shot scenario for novel classes. They normally utilize the information contained in annotated support images of a certain class to perform pixel-wise classification on the query image via non-parametric similarity measurement \cite{Wang_2019_ICCV,liu2020prototype} or parametric decoder \cite{lang2023base,li2021asgnet,zhang2022mfnet}. Current FSS approaches cannot identify base and novel classes simultaneously. 

\paragraph{Generalized Few-Shot Learning.}

Generalized few-shot learning (GFSL) extends FSS by equipping the model with the ability of recognizing the  novel classes with few samples while  preserving the ability of recognizing base classes. For instance, {Kim \etal \cite{kim2023better} propose to apply weight normalization to both base and novel classifiers so as to achieve balanced decision boundaries for both base and novel classes.} Generalized few-shot semantic segmentation (GFSS) is an application of GFSL in semantic segmentation.  
%which addresses the limitation of FSS to segment both base and novel classes in the image. 
GFSS methods like CAPL \cite{tian2022generalized} and PKL \cite{Huang_2023_ICCV} leverage abundant base samples to train base prototypes; subsequently, novel prototypes are generated directly from the support samples of novel classes to work with base prototypes as joint classifiers. On the other hand, approaches such as POP \cite{liu2023learning}, DIaM \cite{Hajimiri_2023_CVPR}, and BCM \cite{sakai2024surprisingly} employ a two-phase training scheme to train both base and novel classifiers. In the pre-training phase, the base classifier is trained with sufficient base samples. In the fine-tuning phase, DIaM \cite{Hajimiri_2023_CVPR} and BCM \cite{sakai2024surprisingly} train the novel classifier using only novel samples, while POP \cite{liu2023learning} uses both base and novel samples.

Following \cite{kim2023better,liu2023learning,Hajimiri_2023_CVPR,sakai2024surprisingly}, our method is built with the two-phase training scheme. We aim to overcome the data imbalance and distribution gap between base and novel classes by effectively transferring the knowledge from base to novel classes, through three new modules, \ie novel prototype modulation, novel classifier calibration, and context consistency learning. 
%The common objective of these three modules is to transfer base knowledge to novel classes during the fine-tuning phase. 
%Notably, unlike POP, we use annotated novel samples and unlabelled base samples to fine-tune the model.}

\section{Method} \label{sec:method}
\subsection{Problem Definition} \label{sec:definition}

For generalized few-shot semantic segmentation, the training data set consists of a base set $D_{base}$ involving $M$ base classes $C_{base}$ with abundant annotated images and a novel set $D_{novel}$ including $N$ novel classes $C_{novel}$ with a few annotated images per class.
%Furthermore, the background category is denoted as $c_0$. 
Note that $C_{base} \cap C_{novel} = \emptyset$. % \cap c_0 = \emptyset$. 

\subsection{Method Overview} \label{sec:method overview}

\begin{figure*}[t]
	\centering
	\includegraphics[width=0.95\textwidth]{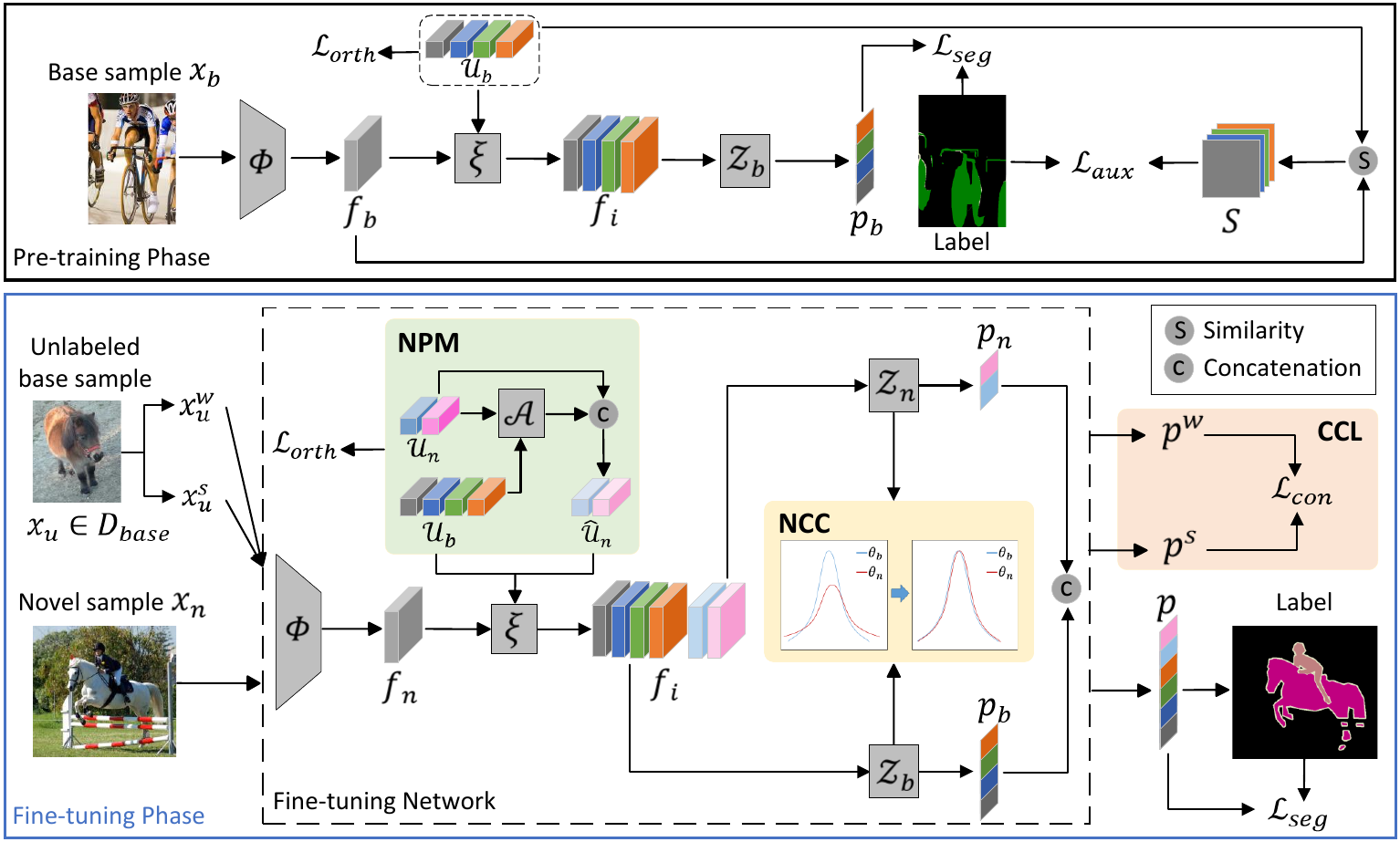}
    %\vspace{-1mm}
	\caption{{Overview of our model. Novel prototype modulation (NPM) module, novel classifier calibration (NCC) module, and context consistency loss (CCL) module are shown in the boxes with green, yellow, and orange backgrounds respectively.}}
        %\vspace{-3mm}
	\label{fig:overview}
\end{figure*}
Figure \ref{fig:overview} illustrates an overview of our proposed method. It follows two-phase framework and involves three new modules, \ie novel prototype modulation (NPM, Sec. \ref{sec:npm}), novel classifier calibration (NCC, Sec. \ref{sec:ncc}), and context consistency learning (CCL, Sec. \ref{sec:ccl}). Specifically, in the pre-training phase, given a base sample $x_b$, we extract its feature $f_b$ using the feature extractor $\Phi$. $f_b$ is then fed into the feature decomposer $\xi$ similar to \cite{liu2023learning}, where $f_b$ is decomposed into $M + 1$ sub-features $\{f_i\}_{i=0}^{M}$ by projecting it onto base prototypes $\mathcal U_b$. 
Each sub-feature $f_i$ is the feature representation for the $i$-th class, and $f_0=f_b-\sum\nolimits_{i=1}^M f_i$ is the background representation. $\mathcal U_b = \{u_i\}_{i=1}^M$ are randomly initialized and updated in the model. Afterward, these sub-features are passed through a learnable base classifier $\mathcal Z_{b}$ to obtain the predicted probability $p_b$ for the per-pixel classification of $x_b$. In the fine-tuning phase, given a novel sample $x_n$, we similarly extract its feature $f_n$ using $\Phi$ and decompose it into $M+N+1$ sub-features $\{f_i\}_{i=0}^{M+N}$ by projecting it onto well-learned base prototypes $\mathcal U_b$ and randomly initialized novel prototypes $\mathcal U_n=\{u_i\}_{i=M+1}^{M+N}$, and $f_0=f_n-\sum\nolimits_{i=1}^{M+N} f_i$. These sub-features are then fed into both the base classifier $\mathcal Z_b$ and the novel classifier $\mathcal Z_n$. Similar to  $\mathcal Z_b$, $\mathcal Z_n$ produces the per-pixel classification result $p_n$ on novel classes; the final predicted probability $p$, %representing the per-pixel classification result for $x_n$, 
is thereby a combination of $p_b$ and $p_n$. Last, two augmented versions (\ie, $x_u^w$ and $x_u^s$) of an unlabeled base sample are processed to compute a consistency loss. Specifically, our proposed modules are all implemented during the fine-tuning phase: 1) NPM is utilized to refine $\mathcal U_n$ by transferring knowledge from $\mathcal U_b$ to $\mathcal U_n$ through the attention mechanism $\mathcal{A}$; 2) NCC is proposed to calibrate $\mathcal Z_n$ according to the weight distributions of $\mathcal Z_b$ and $\mathcal Z_n$; 3) CCL is introduced between $x_u^w$ and $x_u^s$ to reinforce their context consistency and transfer it from base to novel classes.

\subsection{Novel Prototype Modulation}
\label{sec:npm}
Given sufficient base data, the model effectively learns distinct prototypes for base classes; however, it is challenging to learn such novel prototypes given limited novel samples. In fact, novel classes may likely share elemental patterns with base classes \cite{wu2021learning}, we propose to modulate novel prototypes by exploiting their correlations to base prototypes, so as to compensate for the knowledge insufficiency in novel samples.

To capture the correlative information between base and novel classes, we employ the cross-attention mechanism to compute a weighted sum of base prototypes for each novel class. For each novel prototype $u_i \in \mathcal U_n$, it can be reconstructed as a result of the weighted sum of all base prototypes $\mathcal U_b$:
%\vspace{-1mm}
\begin{equation}\label{eq:attention}
%\vspace{-1mm}
\begin{split}
    &\mathcal{A}(u_i, \mathcal U_b) = Softmax(QK^T)V, \\ 
    &Q,K,V = u_iW^Q,\mathcal U_bW^K, \mathcal U_bW^V \\
\end{split}
\end{equation}
where $W^Q$, $W^K$, $W^V$ are three linear layers. $\mathcal{A}(u_i, \mathcal U_b)$ is the reconstructed novel prototype encoded with its correlation to base prototypes. 

Subsequently, we \emph{modulate} the original novel prototype by fusing it with the reconstructed one through concatenation followed by a linear layer. This process generates new prototypes $\hat{\mathcal U}_n$ which replace the original $\mathcal U_n$, preserving the uniqueness of novel classes and their connections to base classes. 
% \begin{equation}\label{eq:fuse}
% \begin{split}
%     \hat{u}_i &= Fuse(u_i,Attention(u_i)) \\
%               &=Linear(concat(u_i,Attention(u_i)))
% \end{split}
% \end{equation}
%This module happens during fine-tuning.
%where $\hat{u}_i$ is a fused novel prototype that integrates the unique feature of the novel class along with the correlative information from base classes.

\subsection{Novel Classifier Calibration} 
\label{sec:ncc}

\begin{figure}[t]
	\centering
	%\vspace{-2mm}
    \includegraphics[width=1.0\columnwidth]{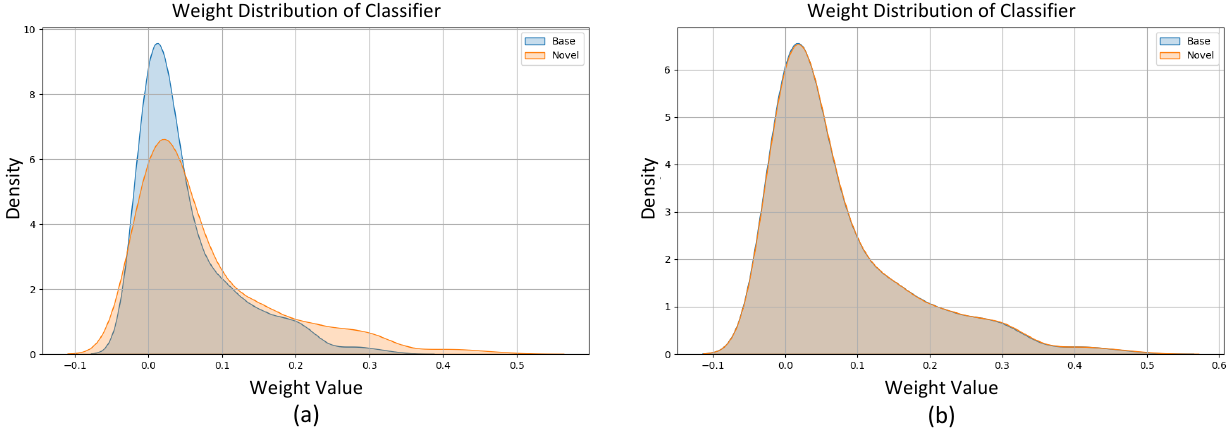}
    %\vspace{-6mm}
    \caption{(a) illustrates the weight distributions of two classifiers without NCC; (b) shows their weight distributions after implementing NCC.}
        %\vspace{-2mm}
	\label{fig:weight}
\end{figure}

When fine-tuning the model on novel classes with only few support samples, the novel classifier is prone to overfitting. In contrast, the base classifier originally trained with sufficient base data is more reliable. The performance on both base and novel classes can be negatively impacted, whilst 
%, as the final prediction is the combination of outputs from base and novel classifiers.
previous approaches often ignore this. We propose to leverage the  knowledge in base classifier to aid in training the novel classifier. 

As reported in \cite{kim2023better}, the position and shape of a decision boundary, which separates different classes in the feature space, are determined by the classifier's weight distribution. Our observation (see Figure \ref{fig:weight}) indicates that the base classifier exhibits a more centric weight distribution; while that of the novel classifier has a larger variance, resulting in potential prediction bias towards it.   
%As proved in \cite{hou2019learning,kim2023better}, the standard deviation of the weight distribution is proportional to the weight norm, and the larger weight norm results in predictions that are more biased toward the corresponding class. 
To address this issue, we calibrate the weight distribution of the novel classifier to align with that of the base classifier. This way can effectively improve the generalization ability of the novel classifier.

We investigate the underlying weight distributions of base and novel classifiers by their means and standard deviations. More specifically, the weights of the base classifier and novel classifier are denoted as  
$\mathcal Z_b = \{\theta_i\}_{i=1}^M \in \mathbb{R}^{d \times M}$ and $\mathcal Z_n = \{\theta_i\}_{i=M+1}^{M+N} \in \mathbb{R}^{d \times N}$, 
respectively; $\theta_i \in \mathbb{R}^d$ represents the weight vector for class $i$. 
%The predicted probability of each sub-feature $f_i$ is computed as: $p_i=\exp(\theta_b^T f_i)/\sum\nolimits_i \exp(\theta_b^T f_i)$.
We calculate the mean and standard deviation for the weight vector in the channel direction:
$\mu_i=\frac{1}{d}\sum\nolimits_{j=1}^d \theta_{ij}$ and $\sigma_i=\sqrt{\frac{1}{d}\sum\nolimits_{j=1}^d(\theta_{ij}-\mu_i)}$. 
%According to these equations, the mean ($\mu_b$) and standard deviation ($\sigma_b$) for the base classifier, as well as the mean ($\mu_n$) and standard deviation ($\sigma_n$) for the novel classifier, can be determined.
Afterward, the average of means and standard deviations over base classes can be given by: $\overline{\mu}_b=\frac{1}{M} \sum\nolimits_{i=1}^{M} \mu_i$ and $\overline{\sigma}_b=\frac{1}{M}\sum_{i=1}^{M}\sigma_i$. Similarly, $\mu_n$, $\sigma_n$, $\overline{\mu}_n$ and $\overline{\sigma}_n$ for the novel classifier $\mathcal Z_n$ can be calculated following the same way. %corresponding to the weights of base and novel classifiers. We compute the average of these vectors along the channel dimension. 

To calibrate the novel classifier, we adjust $\mathcal Z_n$ by first centering it and then shifting it to be in line with $\overline{\mu}_b$:
%to have the same mean as the base classifier during fine-tuning:
\begin{equation}
    \hat{\mathcal Z}_n=\mathcal Z_{n}-\overline{\mu}_{n}+\overline{\mu}_{b}
\end{equation}
where $\overline{\mu}_{n}$ and $\overline{\mu}_{b}$ denote the average of means for novel classes and base classes, respectively; they are replicated for subtraction.  
Secondly, we adjust $\hat{\mathcal Z}_n$ by scaling it according to the ratio of $\overline{\sigma}_b$ to $\sigma_n$:
%scale the standard deviation of the weight distribution of the novel classifier according to the ratio of standard deviation between base and novel classifiers to do what ?. It can be formulated as:
\begin{equation}
    \hat{\mathcal Z}_n = \frac{\overline{\sigma}_{b}}{\sigma_{n}}\hat{\mathcal Z}_n
\end{equation}
%where $\overline{\sigma}_{b}$ denotes the average of standard deviations for base classes. 
%This formulation is utilized to make the weight variance of the novel classifier similar to that of the base classifier. 
The calibration happens during the fine-tuning phase after every epoch. 
%that learns novel classes while retaining the knowledge of base classes.

\subsection{Context Consistency Learning}
\label{sec:ccl}

$K$-shot support samples of novel classes provide very limited contextual information for the class.
%the fine-tuned model with these novel samples is prone to overfitting on novel classes.} 
Nonetheless, base and novel classes may likely share similar contexts. 
%We observe from the dataset that different objects can appear in similar backgrounds, indicating that there is some shared context between base and novel samples. 
%Additionally, fundamental semantic information is often shared between base and novel classes. 
To further leverage base samples for improving the learning of novel classes, we introduce a context consistency learning scheme by incorporating the base samples into the fine-tuning phase following an unsupervised manner. 
%augmenting the base between two versions generated by applying various augmentations to the same sample. 

Directly incorporating the base samples with labels into the fine-tuning phase would bias the model towards base classes.
Inspired by the consistency learning techniques widely used in semi-supervised learning works \cite{sohn2020fixmatch,wei2023segmatch,ouali2020semi}, that is a robust model should yield similar outputs for different perturbed versions of the same image, we introduce a context consistency learning scheme.
As illustrated in Figure \ref{fig:overview} , for a given base image $x_u$, we generate a weakly-augmented version $x_u^w$ using simple augmentation operation, \ie, flipping and cropping, and a strongly-augmented version $x_u^s$ by applying additional augmentation, \ie, cutout, on top of the weakly-augmented version. 

Augmentation like cutout in representative semi-supervised works \cite{sohn2020fixmatch,ouali2020semi} is normally performed randomly on the image without constraints. However, this operation often fails to capture meaningful information. To let the model specifically focus on the contextual part of the image, we instead introduce a context-oriented augmentation operation:   
%We follow specific rules to select the areas for cropping and cutout instead of randomly selecting: 1) We select the background area when cropping. 2) Since background regions typically contain less semantic information, we choose an area close to the objects \textcolor{red}{within the cropped region for} cutout. Specifically, 
%we construct a probability generator to select cropping areas. 
%Let $L$ represent a list of categories of objects present in the image label. For each $l \in L$, we obtain its binary mask $B_l=(M==l)$. From $B_l$, we extract $bboxes_l$, which represent the bounding boxes for objects of class $l$. 
first, we take the bounding box of the ground truth mask for each object class in the image (note that one bounding box is generated when multiple objects of the same class can form a single connected component); second, we randomly select a pixel from the bounding box boundaries of all classes, serving as the center of a $16 \times 16$ squared cutout region. This is because, in the image, pure backgrounds, \eg, grass, and sky, are too simple context for consistency learning, while cutout purely within the objects lacks contextual information for knowledge transfer. Therefore, we select regions close to or with a partial of the objects.  

Let $p^w$ and $p^s$ denote the predicted probabilities output by the segmentation model for $x_u^w$ and $x_u^s$, respectively. We compute the cross-entropy loss between $p^w$ and $p^s$ for consistency as follows:
%\vspace{-2mm}
\begin{equation}
%\vspace{-2mm}
    \mathcal{L}_{con} = \frac{1}{H \times W} \sum_{i=1}^{H \times W} - p_i^w \log(p_i^s)
\end{equation}
where $H \times W$ is the number of pixels in the image.

\subsection{Optimization}
\label{sec:optimization}

\paragraph{Phase 1: Pre-Training.}

In this phase, we only use the base set $D_{base}$ to train the network including the feature extractor $\Phi$, base classifier $\mathcal Z_{b}$ and base prototypes $\mathcal{U}_b$ in the feature decomposer $\xi$. Following previous works \cite{tian2022generalized,liu2023learning,kim2023better}, we calculate the segmentation loss $\mathcal{L}_{seg}$ (a cross-entropy loss) and the orthogonal loss $\mathcal{L}_{orth} =\sum_{i \neq j}|u_i \cdot u_j^T|$ ($u_i, u_j \in \{\mathcal U_b, \mathcal U_n\}$). 

In addition, we introduce an auxiliary loss $\mathcal{L}_{aux}$: we use base prototypes $\mathcal U_b$ as if they were a base classifier to classify the image feature $f_b$ by computing the similarity between $\mathcal U_b$ and $f_b$, and the output result is optimized with the ground truth using the cross-entropy loss (see Figure \ref{fig:overview}). $\mathcal{L}_{aux}$ used in the pre-training phase boosts performance on base classes, as evidenced by the results in both 1-shot and 5-shot settings shown in Table \ref{tab:ablation}.   
The total training loss is: $\mathcal{L}_{total}=\mathcal{L}_{seg}+\mathcal{L}_{orth}+\mathcal{L}_{aux}$. 

\paragraph{Phase 2: Fine-Tuning.}

In this phase, the model is updated with support samples of novel classes alongside unlabeled base samples from $D_{base}$. We first freeze all modules learned in the pre-training phase and then train novel prototypes and novel classifier. The training loss during this phase contains three components, defined as: $\mathcal{L}_{total}=\mathcal{L}_{seg}+\mathcal{L}_{orth}+\mathcal{L}_{con}$, where $\mathcal{L}_{con}$ is the consistency loss shown in Sec. \ref{sec:ccl}.

\section{Experiments} \label{sec:experiments}
\subsection{Datasets and Evaluation Metric} \label{sec:data}

\paragraph{Datasets.}

We evaluate our model on two public datasets, PASCAL-5$^i$ \cite{shaban2017one} and COCO-20$^i$ \cite{nguyen2019feature}. PASCAL-5$^i$ comprises 20 categories and COCO-20$^i$ consists of 80 categories. Object categories in each dataset are evenly split into four folds. Following \cite{tian2022generalized}, we adopt a cross-validation manner to train the model on three folds while testing on one fold. This procedure is repeated four times, and we report the average result. We perform $K \in \{1,5\}$ shot semantic segmentation.

\paragraph{Evaluation Metric.} 
Following previous approaches \cite{tian2022generalized,liu2023learning}, the mean intersection-over-union (mIoU) is adopted as the evaluation metric. $mIoU = \frac{1}{C}\sum\nolimits^{C}_{i=0} IoU_i$, where $C$ is the number of classes,  
%(\textit{e.g.}, $C_{base}=15 +1 %\text{(bakcground)}$ and $C_{novel}=5$ in PASCAL-$5^i$ dataset), 
and $IoU_i$ denotes the intersection-over-union between the predicted segmentation mask and ground truth mask for the $i$-th novel class. To comprehensively evaluate our results, we calculate the average $mIoU$ over the four times validation, denoted by $\overline{mIoU}_{base}$ and $\overline{mIoU}_{novel}$ for the base and novel classes, respectively. Subsequently, we compute the arithmetic mean \cite{liu2023learning} (denoted as ``Mean'') of the $\overline{mIoU}_{base}$ and $\overline{mIoU}_{novel}$. However, the arithmetic mean is dominated by the base classes, which are the majority. To obtain a more balanced metric, we use the harmonic mean \cite{Huang_2023_ICCV} (denoted as ``H-Mean''), formulated as: 
\begin{equation}
    \operatorname{H-Mean}=\frac{2 \cdot \overline{mIoU}_{base} \cdot \overline{mIoU}_{novel}}{\overline{mIoU}_{base}+\overline{mIoU}_{novel}}  
\end{equation}
This approach addresses class unbalance in datasets such as PASCAL-5$^i$ and COCO-20$^i$, where the number of base classes is three times than that of novel classes.

\subsection{Implementation Details} \label{sec:implement}

We follow \cite{liu2023learning} to curate the dataset. 
%This set includes all images from the original training set that contain at least one pixel belonging to base classes for base class pre-training. 
Notably, in the pre-training phase, the pixels of novel classes in base images are treated as background. During the novel class fine-tuning phase, we randomly sample $K \in \{1,5\}$ images from novel classes as support images. All images are cropped to $473 \times 473$ as input to the network during training.
Our method is implemented in PyTorch with NVIDIA A100. We utilize PSPNet \cite{2017pspnet} with ResNet50 \cite{He_2016_CVPR} as the feature extractor. In the pre-training phase, the model is optimized using stochastic gradient descent (SGD) with an initial learning rate of $0.01$, a momentum of $0.9$, and a weight decay of $0.0001$. The model is trained for $50$ epochs on both datasets, and the batch size is set as 8 and 16 for PASCAL-5$^i$ and COCO-20$^i$, respectively. In the fine-tuning phase, we update the model using SGD with a learning rate of $0.01$, training for 500 epochs on both datasets.

% \usepackage{multirow}

\begin{table*}[!t]
% \vspace{-1mm}
\centering
\begin{tabular}{lllllllll}
\hline
Methods & \multicolumn{4}{l}{1-shot}                                        & \multicolumn{4}{l}{5-shot}                                        \\ \cline{2-9}  
        & Base           & Novel          & Mean          & H-Mean          & Base           & Novel          & Mean          & H-Mean          \\ \hline
CAPL \cite{tian2022generalized}    & 65.48          & 18.85          & 54.38          & 29.27          & 66.14          & 22.41          & 55.72          & 33.48          \\
PKL \cite{Huang_2023_ICCV}      & 68.84          & 26.90          & 58.86          & 37.83          & 69.22          & 34.40          & 61.18          & 45.42          \\
DIaM \cite{Hajimiri_2023_CVPR}    & 70.89          & 35.11          & 61.95          & 46.96          & 70.85          & 55.31          & 66.97          & 62.12          \\
POP \cite{liu2023learning}     & 73.92          & 35.51          & 64.77          & 47.97          & 74.78          & 55.87          & 70.27          & 63.96          \\
BCM \cite{sakai2024surprisingly}  & 71.15          & 41.24          & 64.03          & 52.22          & 71.23          & 55.36          & 67.45          & 62.30  \\
Ours    & \textbf{75.23} & \textbf{43.93} & \textbf{67.78} & \textbf{55.47} & \textbf{75.73} & \textbf{57.00} & \textbf{71.28} & \textbf{65.04} \\ \hline
\end{tabular}
%\vspace{-2mm}
\caption{Performance comparison on PASCAL-5$^i$}
\label{tab:comparison_pascal}
\end{table*}

\begin{table*}[!t]
% \vspace{-1mm}
\centering
\begin{tabular}{lllllllll}
\hline
Methods & \multicolumn{4}{l}{1-shot}                                        & \multicolumn{4}{l}{5-shot}     \\ \cline{2-9} 
        & Base           & Novel          & Mean           & H-Mean         & Base  & Novel & Mean  & H-Mean \\ \hline
CAPL \cite{tian2022generalized}    & 44.61          & 7.05           & 35.46          & 12.18          & 45.24 & 11.05 & 36.80 & 17.76  \\
PKL \cite{Huang_2023_ICCV}    & 46.36          & 11.04          & 37.71          & 17.83          & 46.77 & 14.91 & 38.90 & 22.61  \\
DIaM \cite{Hajimiri_2023_CVPR}   & 48.28          & 17.22          & 39.02          & 25.39          & 48.37 & 28.73 & 38.55 & 36.05  \\
POP \cite{liu2023learning}    & 54.71 & 15.31          & 44.98          & 23.92          & 54.90 & 29.97 & 48.75 & 38.77  \\
BCM \cite{sakai2024surprisingly}  & 49.43         & 18.28          & 42.01          & 26.69          & 49.88          & 30.60          & 45.29          & 37.93  \\
Ours    & \textbf{54.81}          & \textbf{21.83} & \textbf{46.96} & \textbf{31.22} & \textbf{55.68}      & \textbf{31.62}     & \textbf{49.95}      & \textbf{40.33}       \\ \hline
\end{tabular}
%\vspace{-2mm}
\caption{Performance comparison on COCO-20$^i$}
\label{tab:comparison_coco}
\end{table*}

\subsection{Performance Comparison} \label{sec:peformance}

\paragraph{Quantitative Analysis.}
\begin{table*}[!t]
% \vspace{-1mm}
\centering
\begin{tabular}{llllllllllll}
\hline
$\mathcal{L}_{aux}$    & NCC   & CCL   & NPM      & \multicolumn{4}{l}{1-shot}               & \multicolumn{4}{l}{5-shot}               \\ \cline{5-12} 
           &            &      &      & Base   & Novel  & Mean   & H-Mean        & Base   & Novel  & Mean   & H-Mean         \\ \hline 
-          & -          & -    & -     & 70.81  & 36.97  & 62.75  & 48.58         & 72.94  & 54.77  & 68.61  & 62.56         \\
\checkmark & -          & -    & -     & 71.40  & 38.82  & 63.64  & 50.29         & 74.06  & 54.62  & 69.44  & 62.87         \\
\checkmark & \checkmark & -    & -     & 75.29  & 39.15  & 66.69  & 51.51         & \textbf{75.74}  & 53.98  & 70.56  & 63.03         \\
\checkmark & \checkmark & \checkmark & - & \textbf{75.40} & 42.49 & 67.56 & 54.82 & 75.68  & 56.24  & 71.05  & 64.53   \\
\checkmark & \checkmark & \checkmark & \checkmark & 75.23  & \textbf{43.93}  & \textbf{67.78}  & \textbf{55.47}         & 75.73  & \textbf{57.00}  & \textbf{71.27}  & \textbf{65.04}         \\ \hline
\end{tabular}
%\vspace{-2mm}
\caption{Ablation study on modules. $\mathcal{L}_{aux}$ represents the auxiliary loss used in base class pre-training; NCC, CCL and NPM denote the novel classifier calibration, context consistency learning and novel prototype modulation, respectively.}
\label{tab:ablation}
\end{table*}

%\vspace{+4mm}
Table \ref{tab:comparison_pascal} and \ref{tab:comparison_coco} present a comparison of several GFSS methods with our method on the PASCAL-5$^i$ and COCO-20$^i$ datasets. Methods such as CAPL \cite{tian2022generalized}, PKL \cite{Huang_2023_ICCV}, DIaM \cite{Hajimiri_2023_CVPR}, and BCM \cite{sakai2024surprisingly} rely solely on novel samples for novel classifier learning, whilst POP \cite{liu2023learning} and our method also use base samples in the learning process. On both datasets, our method outperforms all other methods in both 1-shot and 5-shot scenarios. Specifically, in terms of ``H-Mean'', our approach achieves significant improvements over the previous best, with gains of 3.25\% mIoU and 4.53\% mIoU in the 1-shot setting on PASCAL-5$^i$ and COCO-20$^i$, respectively. Furthermore, in the 5-shot setting, our method surpasses the previous best by over 1\% on both datasets.
%our method surpasses the previous best significantly, \eg in 1-shot by 3.25\% mIoU and 4.53\% mIoU on PASCAL-5$^i$ on COCO-20$^i$, respectively. Additionally, our method improve the previous best over 1\% on both datasets in 5-shot setting.
%The increase on novel classes is especially momentous.
%because our method enhances the knowledge transfer between base and novel classes compared to POP \cite{liu2023learning}. 
Notice that our performance gain diminishes in 5-shot because the complementary knowledge provided by the base classes for novel classes becomes increasingly limited as the number of novel samples increases. These results indicate that our method performs particularly well on novel classes, demonstrating its effectiveness in enhancing knowledge transfer from base to novel classes for GFSS.  

\begin{figure}[!t] 
    \centering
    \includegraphics[width=1.0\columnwidth]{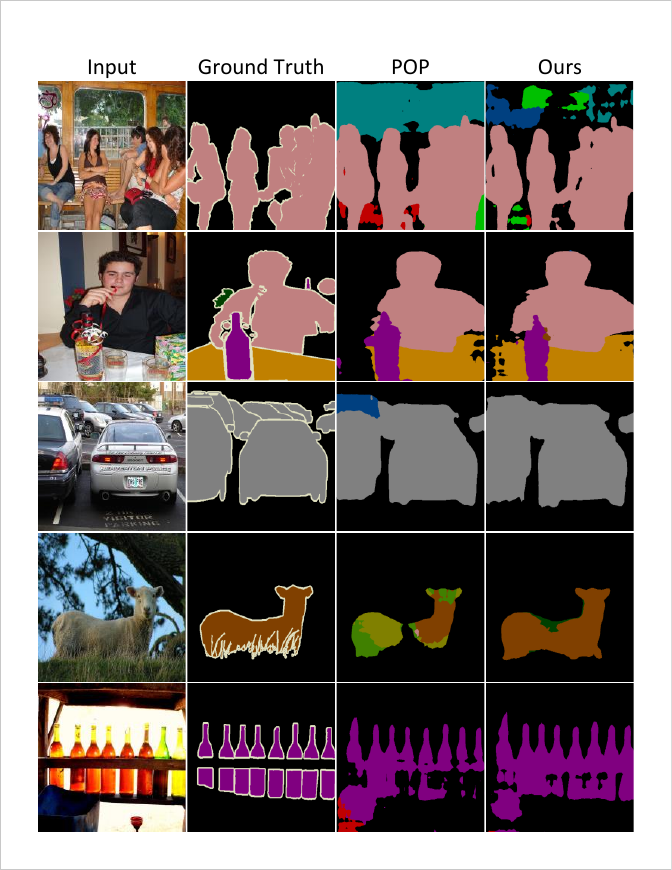}
    \caption{Qualitative results of our method and POP \cite{liu2023learning} on PASCAL-5$^i$.}
    \label{fig:seg}
    %\vspace{-2mm}
\end{figure}

\paragraph{Qualitative Comparison.}
Figure \ref{fig:seg} illustrates five examples from PASCAL-5$^i$ using our method and POP \cite{liu2023learning} in the 1-shot scenario. Our method shows superior segmentation performance. For instance, the first example in the first row demonstrates the effectiveness of our method in reducing image noise; the last example indicates that our method improves the model's ability to segment different parts of objects.

\subsection{Ablation Studies} \label{sec:ablation}

In this section, we investigate the effectiveness of each module proposed in our method by conducting a series of ablation studies on PASCAL-5$^i$. 

\paragraph{Novel Prototype Modulation (NPM).}

\begin{table}[!t]
\centering
    % \vspace{-2mm}
    \begin{tabular}{lllll}
        \hline
        Methods & \multicolumn{4}{l}{1-shot} \\
        \cline{2-5}
        ~ & Base & Novel & Mean & H-Mean \\
        \hline
        CS & \textbf{75.44} & 41.32 & 67.32 & 53.39 \\
        CA & 75.23 & \textbf{43.93} & \textbf{67.78} & \textbf{55.47} \\
        \hline
    \end{tabular}
    %\vspace{-2mm}
    \caption{Ablation study on the NPM. CS denotes cosine similarity, while CA denotes cross attention.}
    \label{tab:npm}
%\end{minipage}
\end{table}
When removing NPM, a decrease of 1.44\% mIoU on novel classes can be observed in Table~\ref{tab:ablation}, which indicates that this module effectively transfers correlation information from base to novel prototypes.
To investigate NPM, we compare NPM configured with different reconstruction methods. The results in Table~\ref{tab:npm} demonstrate that using cross-attention (CA) to compute the correlation between base and novel prototypes is superior to using cosine similarity (CS).
We attribute this to the strong ability of cross-attention to capture correlations between base and novel classes, thereby enhancing knowledge transfer from base to novel classes.

\paragraph{Novel Classifier Calibration (NCC).}

\begin{table}[!t]
%\begin{minipage}{0.49\linewidth}
\centering
% \vspace{-2mm}
    \begin{tabular}{lllll}
        \hline
        Methods & \multicolumn{4}{l}{1-shot} \\
        \cline{2-5}
        ~ & Base & Novel & Mean & H-Mean \\
        \hline
        w/o NCC & 71.40 & 38.82 & 63.64 & 50.29 \\
        w/ NBCC & 72.61 & 38.34 & 64.56 & 50.18 \\
        w/ NCC & \textbf{75.29} & \textbf{39.15} & \textbf{66.69} & \textbf{51.51} \\
        \hline
    \end{tabular}
    %\vspace{-2mm}
    \caption{Ablation study on NCC. NBCC denotes calibrating both base and novel classifiers.}
    \label{tab:ca}
%\end{minipage}
\end{table}
Table \ref{tab:ablation} presents a 3.89\% mIoU decrease on base classes by removing NCC and a slight drop, 0.33\% mIoU, on novel classes, which demonstrates that it is effective to calibrate the weight distribution of the novel classifier, reducing its bias.
To further investigate the method of novel classifier calibration, we investigate another variant such as 
% oppositely calibrating the weight distribution of the base classifier according to that of the novel classifier (denoted as BCC)
calibrating both base and novel classifiers according to the averaged statistics of their weight distributions (NBCC).
Table~\ref{tab:ca} shows that both NCC and NBCC enhance performance on base classes. However, NCC achieves more significant improvements, with a 2.68$\%$ mIoU increase on base classes and a 0.81$\%$ mIoU increase on novel classes compared to NBCC, adjusting the weight distribution of base classifier somewhat disrupts the well-learned base classifier.

%Table \ref{tab:ablation} presents the experimental results of an ablation study on different modules in our method. Our model with the combination of all three modules achieves the best performance, with H-Mean of 55.47$\%$ and 65.04 $\%$ in 1-shot and 5-shot settings, respectively. Analyzing 1-shot H-Mean, we observe a decrease of approximately 1.5 $\%$ when ablating NPM, a 2.2$\%$ reduction by removing CC and a 1.2$\%$ drop by removing NCC. Notably, NCC primarily benefits the base classes since it calibrates the novel classifier according to the base classifier. 
%which effectively preventing the fine-tuned model from biasing towards novel classes. 
%CC and NPM enhance performance on novel classes without causing forgetting on base classes as both of them transfer effective information from base to novel classes. Additionally, AL, proposed as an auxiliary loss, boosts performance on base classes, as evidenced by the results in both 1-shot and 5-shot settings.

%Table \ref{tab:ablation} shows that there is a NCC module

\paragraph{Context Consistency Learning (CCL).}

\begin{table}[!t]
\centering
% \vspace{-2mm}
    % \vspace{+1px}
    \begin{tabular}{lllll}
        \hline
        Methods & \multicolumn{4}{l}{1-shot} \\
        \cline{2-5}
        ~ & Base & Novel & Mean & H-Mean \\
        \hline 
        w/ label & 74.61 & 39.05 & 66.14 & 51.27 \\
        w/o label & \textbf{75.40} & \textbf{42.49} & \textbf{67.56} & \textbf{54.82} \\
        \hline
    \end{tabular}
    %\vspace{-2mm}
\caption{Ablation study on the CCL. Comparison between using or not using labels of base sample in fine-tuning.}
\label{tab:cr}
\end{table}

\begin{table}[!t]
\centering
    % \vspace{-2mm}
    \begin{tabular}{lllll}
        \hline
        Methods & \multicolumn{4}{l}{1-shot} \\
        \cline{2-5}
        ~ & Base & Novel & Mean & H-Mean \\
        \hline
        base $+$ novel & \textbf{75.40} & \textbf{42.49} & \textbf{67.56} & \textbf{54.82} \\
        % \hdashline
        w/o base & 75.12 & 36.58 & 65.94 & 49.20 \\
        w/o novel & 75.04 & 35.30 & 65.57 & 48.01 \\
        \hline
    \end{tabular}
    %\vspace{-2mm}
    \caption{Ablation study on the CCL. Comparison between different classifiers.}
    \label{tab:classifier}
\end{table}

\begin{table}[!t]
\centering
    % \vspace{-2mm}
    \begin{tabular}{lllll}
        \hline
        Methods & \multicolumn{4}{l}{1-shot} \\
        \cline{2-5}
        ~ & Base & Novel & Mean & H-Mean \\
        \hline
        only novel & 75.39 & 39.28 & 66.79 & 51.65 \\
        only base & 75.40 & \textbf{42.49} & \textbf{67.56} & \textbf{54.82} \\
        base $+$ novel & \textbf{75.51} & 41.20 & 67.43 & 53.31 \\
        \hline
    \end{tabular}
    % \vspace{-1mm}
    \caption{Ablation study on the CCL. Comparison between different sources of samples.}
    \label{tab:sample}
\end{table}

\begin{table}[!t]
\centering
    % \vspace{-2mm}
    \begin{tabular}{lllll}
        \hline
        Methods & \multicolumn{4}{l}{1-shot} \\
        \cline{2-5}
        ~ & Base & Novel & Mean & H-Mean \\
        \hline
        Randaugment & 73.06 & 35.91 & 64.36 & 48.15 \\
        cutout & \textbf{75.40} & \textbf{42.49} & \textbf{67.56} & \textbf{54.82} \\
        \hline
    \end{tabular}
    % \vspace{-1mm}
    \caption{Ablation study on the CCL. Comparison between different augmentation techniques.}
    \label{tab:augment}
\end{table}

\begin{table}[!t]
\centering
    % \vspace{-2mm}
    \begin{tabular}{lllll}
        \hline
        Methods & \multicolumn{4}{l}{1-shot} \\
        \cline{2-5}
        ~ & Base & Novel & Mean & H-Mean \\
        \hline
        wcutout & 74.98 & 40.34 & 66.73 & 52.46\\
        ocutout & 75.17 & 40.62 & 66.94 & 52.74 \\
        icutout & 75.07 & 40.46 & 66.83  & 52.58\\
        bcutout (ours) & \textbf{75.40} & \textbf{42.49} & \textbf{67.56} & \textbf{54.82} \\
        \hline
    \end{tabular}
    % \vspace{-1mm}
    \caption{Ablation study on the CCL. Comparison between different cutout methods.}
    \label{tab:cutout}
\end{table}
In Table~\ref{tab:ablation}, there is a significant drop (\ie, -3.34\% mIoU) on novel classes when ablating CCL, which shows the effectiveness of this scheme.
To further validate it, we conduct several experiments. In Table~\ref{tab:cr}, we compare the experimental results of the fine-tuning phase using labeled or unlabeled base samples.
The results indicate that the label-free method (w/o label) outperforms the labeled one (w/ label), showing improvements in both base and novel classes.
This suggests that using explicit supervision for base classes during fine-tuning increases the imbalance between base and novel classes, resulting in the model biased towards base classes.

To evaluate the necessity of using both base and novel classifiers when applying consistency to base samples, we conduct an experiment that removes the base or novel classifier respectively.
% Table \ref{tab:classifier} demonstrates that both classifiers play a crucial role in the final prediction.
In Table~\ref{tab:classifier}, we observe that omitting either the base classifier (w/o base) or the novel classifier (w/o novel), as compared to employing both (base $+$ novel), results in a significant performance decline, which demonstrates that both the base and novel classifiers play crucial roles in the final prediction.

Furthermore, Table~\ref{tab:sample} shows the results of performing consistency learning on samples from novel/base set.
The results show that using only novel samples (only novel) or using both base and novel samples (base $+$ novel) is inferior to using only base samples (only base).
This is likely because novel samples are fully utilized through the supervised learning, incorporating them also in the unsupervised learning might confuse the model.  
%and 2) the context of novel samples is largely encompassed within base samples. 

Moreover, we study the augmentation operations for context consistency learning.
First, regarding the strong augmentation, instead of choosing cutout, we follow the method Randaugment~\cite{cubuk2020randaugment} to use photometric transformations (\eg, contrast, brightness, and sharpness).
%by conducting an experiment to evaluate the effect of different augmentations on context consistency. 
As shown in Table~\ref{tab:augment}, a significant performance decrease is observed when using Randaugment. 
Second, we study the proposed context-oriented augmentation operation by comparing different cutout methods. Our proposed cutout is performed by randomly selecting a pixel from the bounding box boundaries to serve as the center of the cutout region (bcutout). For comparison, we consider three alternative methods: wcutout, where the cutout region is randomly selected within the bounding boxes; ocutout, where the cutout region is randomly selected outside the bounding boxes; icutout, where the cutout region is selected randomly from the entire image. 

As shown in Table~\ref{tab:cutout}, the comparison between our proposed bcutout with the wcutout, ocutout, and icutout indicates that our bcutout outperforms the others, as the region near the bounding box boundaries contains richer contextual knowledge.

\subsection{Discussion and Visualization}

\paragraph{Class Split Analysis.}
\begin{table}[!t]
\centering
    % \vspace{-2mm}
    \begin{tabular}{lllll}
        \hline
        Methods & \multicolumn{4}{l}{1-shot} \\
        \cline{2-5}
        ~ & Base & Novel & Mean & H-Mean \\
        \hline
        Baseline &85.36 & 21.74 & 70.21 & 34.65 \\
        Ours & \textbf{86.96} & \textbf{24.39} & \textbf{72.06} & \textbf{38.10} \\
        \hline
    \end{tabular}
    % \vspace{-1mm}
    \caption{Ablation study on a challenging split.}
    \label{tab:split}
\end{table}
To evaluate the effectiveness of our method across various splits, we conduct experiments using a challenging split of PASCAL-5i with \emph{aeroplane}, \emph{bicycle}, \emph{boat}, \emph{bus}, \emph{car}, \emph{motorbike}, and \emph{train} being novel classes, and \emph{bird}, \emph{cat}, \emph{cow}, \emph{dog}, \emph{horse}, \emph{person}, and \emph{sheep} being base classes. The experiment is conducted in the 1-shot setting and the mIoU is reported in Table \ref{tab:split}.
The results show that our method improves both base and novel class performance over POP \cite{liu2023learning}, which serves as the baseline of our method. This demonstrates that our method is effective even when base and novel classes are not intuitively correlated. This effectiveness can be attributed to the presence of shared elemental patterns in the low-level visual features (e.g. texture, shape) between base and novel classes. Furthermore, from a biomimicry perspective, the structures and functions of many man-made objects are indeed inspired by nature. For example, the connections between planes and birds, cars and horses.

\begin{figure}[!t] 
    \centering
    \includegraphics[width=1.0\columnwidth]{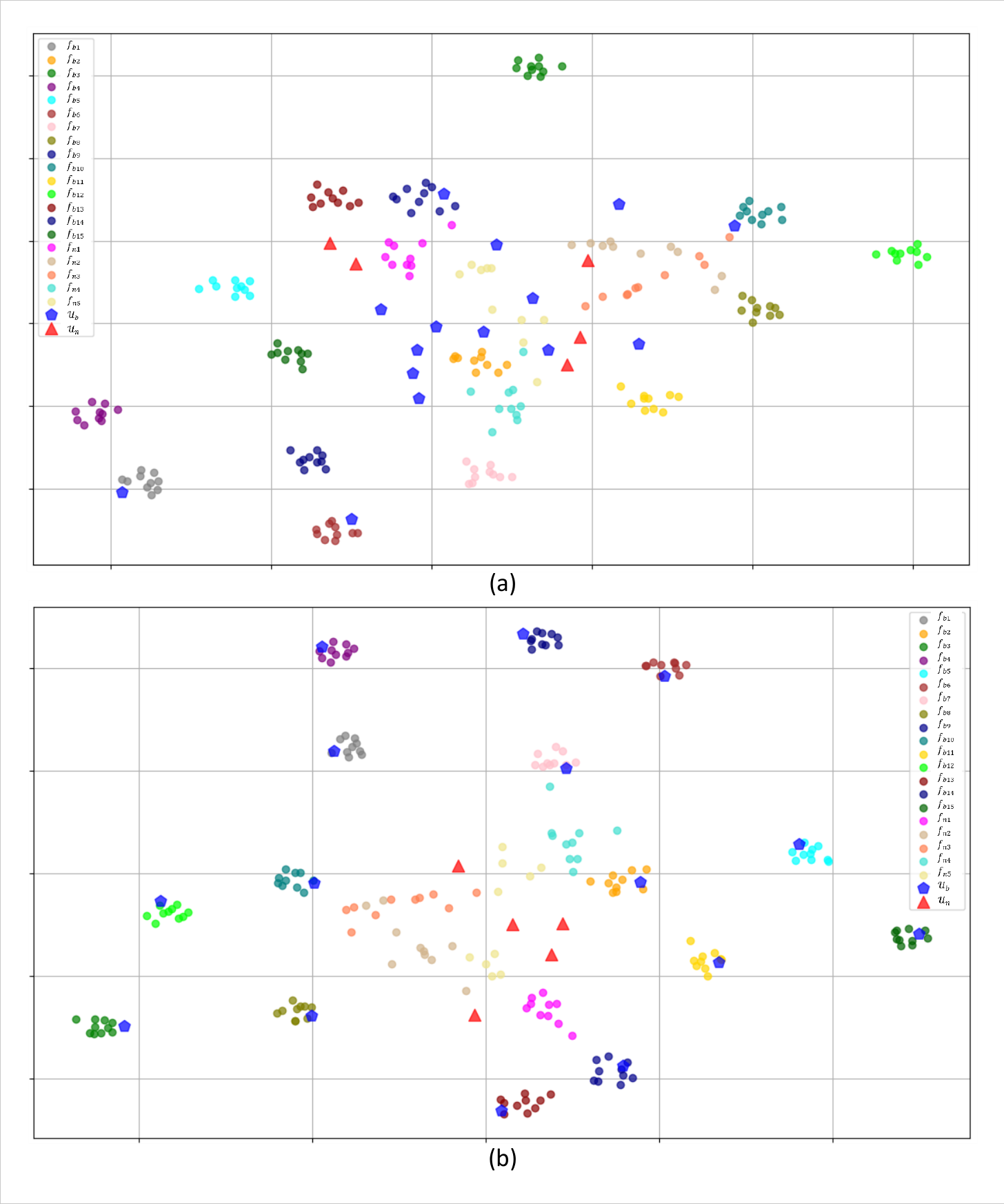}
    \caption{t-SNE visualization of features and class prototypes. (a) illustrates the t-SNE visualization of POP \cite{liu2023learning}; (b) shows the t-SNE visualization of our proposed method. The base prototypes ($\mathcal{U}_b$) are represented as blue pentagrams, and the novel prototypes ($\mathcal{U}_n$) are shown as red rectangles. The features of base classes ($\mathcal{F}_b$) and novel classes ($\mathcal{F}_n$) are visualized as dots, where features of the same class are clustered and represented by dots of the same color.}
    \label{fig:tsne}
\end{figure}

\paragraph{t-SNE Visualization.}
We utilize t-SNE to visualize all base and novel prototypes, along with the features of 20 samples per class. In Figure \ref{fig:tsne} (a), the t-SNE visualization of POP \cite{liu2023learning} is presented, where $\mathcal U_b$ and $\mathcal U_n$ denote base prototypes and novel prototypes, respectively, while $\mathcal{F}_b$ and $\mathcal{F}_n$ represent the features of base and novel classes. Similarly, Figure \ref{fig:tsne} (b) illustrates the t-SNE visualization of our proposed method using the same symbols. Despite both methods extracting features from the same structure with the same parameters, the t-SNE visualizations show that base and novel prototypes become more representative after applying our proposed method.

\section{Conclusion}
%\paragraph{Conclusion.}

In this work, we propose to enhance generalized few-shot semantic segmentation through effective knowledge transfer. During the fine-tuning stage, we design three modules to facilitate knowledge transfer from base to novel classes. First, the novel prototype modulation module is proposed to adjust the novel prototypes by exploiting their correlations with base prototypes. Second, the novel classifier calibration module aims to calibrate the weight distribution of the novel classifier using mean shifting and standard deviation scaling according to that of the base classifier. Last, to further make use of the contextual knowledge in base classes, we introduce a context consistency learning scheme to transfer the contextual information from base to novel classes. We conduct extensive experiments on PASCAL-5$^i$ and COCO-20$^i$, which demonstrate that our method effectively improves previous GFSS approaches. Future work will focus on leveraging text information via multimodal large language model into this framework. 

\section*{Acknowledgments}

This work is supported by the Fundamental Research Funds for the Central Universities. Xinyue Chen is funded by the China Scholarship Council.

\bibliography{aaai25}

\begin{thebibliography}{34}
\providecommand{\natexlab}[1]{#1}

\bibitem[{Chen et~al.(2017)Chen, Papandreou, Kokkinos, Murphy, and Yuille}]{chen2017deeplab}
Chen, L.-C.; Papandreou, G.; Kokkinos, I.; Murphy, K.; and Yuille, A.~L. 2017.
\newblock Deeplab: Semantic image segmentation with deep convolutional nets, atrous convolution, and fully connected crfs.
\newblock \emph{IEEE transactions on pattern analysis and machine intelligence}, 40(4): 834--848.

\bibitem[{Chen and Shi(2024)}]{chen2024memory}
Chen, X.; and Shi, M. 2024.
\newblock Memory-guided Network with Uncertainty-based Feature Augmentation for Few-shot Semantic Segmentation.
\newblock In \emph{IEEE International Conference on Multimedia and Expo (ICME)}.

\bibitem[{Chen et~al.(2023)Chen, Wang, Xu, and Shi}]{chen2023distilling}
Chen, X.; Wang, Y.; Xu, Y.; and Shi, M. 2023.
\newblock Distilling base-and-meta network with contrastive learning for few-shot semantic segmentation.
\newblock \emph{Autonomous Intelligent Systems}, 3(1): 11.

\bibitem[{Cheng et~al.(2022)Cheng, Misra, Schwing, Kirillov, and Girdhar}]{2022mask2former}
Cheng, B.; Misra, I.; Schwing, A.~G.; Kirillov, A.; and Girdhar, R. 2022.
\newblock Masked-attention mask transformer for universal image segmentation.
\newblock In \emph{Proceedings of the IEEE/CVF Conference on Computer Vision and Pattern Recognition}, 1290--1299.

\bibitem[{Cubuk et~al.(2020)Cubuk, Zoph, Shlens, and Le}]{cubuk2020randaugment}
Cubuk, E.~D.; Zoph, B.; Shlens, J.; and Le, Q.~V. 2020.
\newblock Randaugment: Practical automated data augmentation with a reduced search space.
\newblock In \emph{Proceedings of the IEEE/CVF conference on computer vision and pattern recognition workshops}, 702--703.

\bibitem[{Dosovitskiy et~al.(2020)Dosovitskiy, Beyer, Kolesnikov, Weissenborn, Zhai, Unterthiner, Dehghani, Minderer, Heigold, Gelly et~al.}]{dosovitskiy2020vit}
Dosovitskiy, A.; Beyer, L.; Kolesnikov, A.; Weissenborn, D.; Zhai, X.; Unterthiner, T.; Dehghani, M.; Minderer, M.; Heigold, G.; Gelly, S.; et~al. 2020.
\newblock An image is worth 16x16 words: Transformers for image recognition at scale.
\newblock \emph{arXiv preprint arXiv:2010.11929}.

\bibitem[{Hajimiri et~al.(2023)Hajimiri, Boudiaf, Ben~Ayed, and Dolz}]{Hajimiri_2023_CVPR}
Hajimiri, S.; Boudiaf, M.; Ben~Ayed, I.; and Dolz, J. 2023.
\newblock A Strong Baseline for Generalized Few-Shot Semantic Segmentation.
\newblock In \emph{Proceedings of the IEEE/CVF Conference on Computer Vision and Pattern Recognition (CVPR)}, 11269--11278.

\bibitem[{He et~al.(2016)He, Zhang, Ren, and Sun}]{He_2016_CVPR}
He, K.; Zhang, X.; Ren, S.; and Sun, J. 2016.
\newblock Deep residual learning for image recognition.
\newblock In \emph{Proceedings of the IEEE conference on computer vision and pattern recognition}, 770--778.

\bibitem[{Hu, Sun, and Yang(2022)}]{hu2022suppressing}
Hu, Z.; Sun, Y.; and Yang, Y. 2022.
\newblock Suppressing the heterogeneity: A strong feature extractor for few-shot segmentation.
\newblock In \emph{The Eleventh International Conference on Learning Representations}.

\bibitem[{Huang et~al.(2023)Huang, Wang, Xi, and Gao}]{Huang_2023_ICCV}
Huang, K.; Wang, F.; Xi, Y.; and Gao, Y. 2023.
\newblock Prototypical Kernel Learning and Open-set Foreground Perception for Generalized Few-shot Semantic Segmentation.
\newblock In \emph{Proceedings of the IEEE/CVF International Conference on Computer Vision (ICCV)}, 19256--19265.

\bibitem[{Kim and Choi(2023)}]{kim2023better}
Kim, S.-W.; and Choi, D.-W. 2023.
\newblock Better generalized few-shot learning even without base data.
\newblock In \emph{Proceedings of the AAAI Conference on Artificial Intelligence}, volume~37, 8282--8290.

\bibitem[{Lang et~al.(2023)Lang, Cheng, Tu, Li, and Han}]{lang2023base}
Lang, C.; Cheng, G.; Tu, B.; Li, C.; and Han, J. 2023.
\newblock Base and meta: A new perspective on few-shot segmentation.
\newblock \emph{IEEE Transactions on Pattern Analysis and Machine Intelligence}, 45(9): 10669--10686.

\bibitem[{Li et~al.(2021)Li, Jampani, Sevilla-Lara, Sun, Kim, and Kim}]{li2021asgnet}
Li, G.; Jampani, V.; Sevilla-Lara, L.; Sun, D.; Kim, J.; and Kim, J. 2021.
\newblock Adaptive prototype learning and allocation for few-shot segmentation.
\newblock In \emph{Proceedings of the IEEE/CVF conference on computer vision and pattern recognition}, 8334--8343.

\bibitem[{Li et~al.(2022)Li, Yao, Pan, and Mei}]{li2022contextual}
Li, Y.; Yao, T.; Pan, Y.; and Mei, T. 2022.
\newblock Contextual transformer networks for visual recognition.
\newblock \emph{IEEE Transactions on Pattern Analysis and Machine Intelligence}, 45(2): 1489--1500.

\bibitem[{Liu and Qin(2020)}]{liu2020prototype}
Liu, J.; and Qin, Y. 2020.
\newblock Prototype refinement network for few-shot segmentation.
\newblock \emph{arXiv preprint arXiv:2002.03579}.

\bibitem[{Liu et~al.(2023)Liu, Zhang, Qiu, Xie, Zhang, and Yao}]{liu2023learning}
Liu, S.-A.; Zhang, Y.; Qiu, Z.; Xie, H.; Zhang, Y.; and Yao, T. 2023.
\newblock Learning orthogonal prototypes for generalized few-shot semantic segmentation.
\newblock In \emph{Proceedings of the IEEE/CVF Conference on Computer Vision and Pattern Recognition}, 11319--11328.

\bibitem[{Long, Shelhamer, and Darrell(2015)}]{long2015fcn}
Long, J.; Shelhamer, E.; and Darrell, T. 2015.
\newblock Fully convolutional networks for semantic segmentation.
\newblock In \emph{Proceedings of the IEEE conference on computer vision and pattern recognition}, 3431--3440.

\bibitem[{Lu et~al.(2023)Lu, He, Li, Song, and Xiang}]{lu2023prediction}
Lu, Z.; He, S.; Li, D.; Song, Y.-Z.; and Xiang, T. 2023.
\newblock Prediction calibration for generalized few-shot semantic segmentation.
\newblock \emph{IEEE transactions on image processing}, 32: 3311--3323.

\bibitem[{Lu et~al.(2021)Lu, He, Zhu, Zhang, Song, and Xiang}]{lu2021simpler}
Lu, Z.; He, S.; Zhu, X.; Zhang, L.; Song, Y.-Z.; and Xiang, T. 2021.
\newblock Simpler is better: Few-shot semantic segmentation with classifier weight transformer.
\newblock In \emph{Proceedings of the IEEE/CVF International Conference on Computer Vision}, 8741--8750.

\bibitem[{Nguyen and Todorovic(2019)}]{nguyen2019feature}
Nguyen, K.; and Todorovic, S. 2019.
\newblock Feature weighting and boosting for few-shot segmentation.
\newblock In \emph{Proceedings of the IEEE/CVF International Conference on Computer Vision}, 622--631.

\bibitem[{Ouali, Hudelot, and Tami(2020)}]{ouali2020semi}
Ouali, Y.; Hudelot, C.; and Tami, M. 2020.
\newblock Semi-supervised semantic segmentation with cross-consistency training.
\newblock In \emph{Proceedings of the IEEE/CVF conference on computer vision and pattern recognition}, 12674--12684.

\bibitem[{Sakai et~al.(2024)Sakai, Qiu, Katsuki, Kimura, Osogami, and Inoue}]{sakai2024surprisingly}
Sakai, T.; Qiu, H.; Katsuki, T.; Kimura, D.; Osogami, T.; and Inoue, T. 2024.
\newblock A Surprisingly Simple Approach to Generalized Few-Shot Semantic Segmentation.
\newblock In \emph{The Thirty-eighth Annual Conference on Neural Information Processing Systems}.

\bibitem[{Shaban et~al.(2017)Shaban, Bansal, Liu, Essa, and Boots}]{shaban2017one}
Shaban, A.; Bansal, S.; Liu, Z.; Essa, I.; and Boots, B. 2017.
\newblock One-shot learning for semantic segmentation.
\newblock \emph{arXiv preprint arXiv:1709.03410}.

\bibitem[{Sohn et~al.(2020)Sohn, Berthelot, Carlini, Zhang, Zhang, Raffel, Cubuk, Kurakin, and Li}]{sohn2020fixmatch}
Sohn, K.; Berthelot, D.; Carlini, N.; Zhang, Z.; Zhang, H.; Raffel, C.~A.; Cubuk, E.~D.; Kurakin, A.; and Li, C.-L. 2020.
\newblock Fixmatch: Simplifying semi-supervised learning with consistency and confidence.
\newblock \emph{Advances in neural information processing systems}, 33: 596--608.

\bibitem[{Tian et~al.(2022)Tian, Lai, Jiang, Liu, Shu, Zhao, and Jia}]{tian2022generalized}
Tian, Z.; Lai, X.; Jiang, L.; Liu, S.; Shu, M.; Zhao, H.; and Jia, J. 2022.
\newblock Generalized few-shot semantic segmentation.
\newblock In \emph{Proceedings of the IEEE/CVF Conference on Computer Vision and Pattern Recognition}, 11563--11572.

\bibitem[{Wang et~al.(2019)Wang, Liew, Zou, Zhou, and Feng}]{Wang_2019_ICCV}
Wang, K.; Liew, J.~H.; Zou, Y.; Zhou, D.; and Feng, J. 2019.
\newblock Panet: Few-shot image semantic segmentation with prototype alignment.
\newblock In \emph{proceedings of the IEEE/CVF international conference on computer vision}, 9197--9206.

\bibitem[{Wei et~al.(2023)Wei, Budd, Garcia-Peraza-Herrera, Dorent, Shi, and Vercauteren}]{wei2023segmatch}
Wei, M.; Budd, C.; Garcia-Peraza-Herrera, L.~C.; Dorent, R.; Shi, M.; and Vercauteren, T. 2023.
\newblock SegMatch: A semi-supervised learning method for surgical instrument segmentation.
\newblock \emph{arXiv preprint arXiv:2308.05232}.

\bibitem[{Wu et~al.(2021)Wu, Shi, Lin, and Cai}]{wu2021learning}
Wu, Z.; Shi, X.; Lin, G.; and Cai, J. 2021.
\newblock Learning meta-class memory for few-shot semantic segmentation.
\newblock In \emph{Proceedings of the IEEE/CVF International Conference on Computer Vision}, 517--526.

\bibitem[{Yu and Koltun(2015)}]{yu2015multi}
Yu, F.; and Koltun, V. 2015.
\newblock Multi-scale context aggregation by dilated convolutions.
\newblock \emph{arXiv preprint arXiv:1511.07122}.

\bibitem[{Zhang et~al.(2021)Zhang, Kang, Yang, and Wei}]{zhang2021cyc}
Zhang, G.; Kang, G.; Yang, Y.; and Wei, Y. 2021.
\newblock Few-shot segmentation via cycle-consistent transformer.
\newblock \emph{Advances in Neural Information Processing Systems}, 34: 21984--21996.

\bibitem[{Zhang et~al.(2022)Zhang, Navasardyan, Chen, Zhao, Wei, Shi et~al.}]{zhang2022mm-former}
Zhang, G.; Navasardyan, S.; Chen, L.; Zhao, Y.; Wei, Y.; Shi, H.; et~al. 2022.
\newblock Mask matching transformer for few-shot segmentation.
\newblock \emph{Advances in Neural Information Processing Systems}, 35: 823--836.

\bibitem[{Zhang, Shi, and Li(2022)}]{zhang2022mfnet}
Zhang, M.; Shi, M.; and Li, L. 2022.
\newblock MFNet: Multiclass few-shot segmentation network with pixel-wise metric learning.
\newblock \emph{IEEE Transactions on Circuits and Systems for Video Technology}, 32(12): 8586--8598.

\bibitem[{Zhang et~al.(2024)Zhang, Shi, Su, and Wang}]{zhang2024memory}
Zhang, Y.; Shi, M.; Su, T.; and Wang, H. 2024.
\newblock Memory-Based Contrastive Learning with Optimized Sampling for Incremental Few-Shot Semantic Segmentation.
\newblock In \emph{IEEE International Symposium on Circuits and Systems (ISCAS)}, 1--5. IEEE.

\bibitem[{Zhao et~al.(2017)Zhao, Shi, Qi, Wang, and Jia}]{2017pspnet}
Zhao, H.; Shi, J.; Qi, X.; Wang, X.; and Jia, J. 2017.
\newblock Pyramid scene parsing network.
\newblock In \emph{Proceedings of the IEEE conference on computer vision and pattern recognition}, 2881--2890.

\end{thebibliography}

\end{document}